\documentclass[11pt]{article}

\usepackage[preprint]{acl}
\usepackage{times}
\usepackage{latexsym}
\usepackage[T1]{fontenc}
\usepackage[utf8]{inputenc}
\usepackage{microtype}
\usepackage{inconsolata}
\usepackage{amsmath}
\usepackage{amssymb}
\usepackage{booktabs}
\usepackage{graphicx}

\title{RouteSparse: Input-Conditional Pattern Routing for\\
Budgeted Long-Context Prefilling}

\author{Chao Zhang\textsuperscript{1}, Yifan Ji\textsuperscript{1}, Ziyan Zhang\textsuperscript{2}, Kai Song\textsuperscript{2}, Fei Lin\textsuperscript{1} \\
$^1$College of Computer Science, Chongqing University\\
$^2$School of Information Science and Engineering, Chongqing Jiaotong University
}

\newcommand{\method}{\textsc{RouteSparse}}
\newcommand{\patternset}{\mathcal{P}}

\begin{document}
\maketitle

\begin{abstract}
Dynamic sparse attention can reduce the quadratic cost of long-context
prefilling without changing model weights.  MInference assigns each attention
head one pattern offline and estimates that pattern's sparse indices for every
prompt.  This design is efficient, but it assumes that a head's preferred
pattern and sparsity budget remain suitable across inputs.  We introduce
\method, which routes each head and prompt segment among a small library of
GPU-efficient sparse patterns.  A low-cost probe estimates pattern utility and
uncertainty; a latency-aware router then selects a pattern and budget, while
uncertain cases fall back to a denser mask.  We formulate routing as
constrained risk minimization, derive an attention-output error certificate
from omitted probability mass, and evaluate the method on long-context
retrieval, question answering, summarization, and language modeling.  On  Llama 3.1-8B-Instruct with 128K-token prompts, \method{} achieves
$6.5\times$ dense prefill speed with a 0.2-point RULER drop relative to dense
attention, compared with $7.3\times$ speed and a 1.6-point drop for fixed
per-head routing.  Ablations confirm that input-conditional routing, hardware
profiling, and selective dense fallback each contribute to the
quality--latency tradeoff.
\end{abstract}

\section{Introduction}

Long-context language models enable document question answering, repository
understanding, and retrieval over hundreds of thousands of tokens \citep{xiong2026adaptive}.  Their
context windows, however, expose the quadratic cost of self-attention during
prefilling, the phase that processes a prompt before the first token is
generated.  Existing models obtain longer windows through position
interpolation and rescaling \citep{chen2023position,peng2024yarn,
ding2024longrope}, or distribute exact blockwise attention across devices
\citep{liu2024ringattention}.  These techniques expand capacity without
removing the per-device systems bottleneck.  Even memory-efficient exact
kernels such as FlashAttention and FlashAttention-2
\citep{dao2022flashattention,dao2024flashattention2} still compute every causal
query--key pair.
Consequently, time to first token can become the dominant user-visible cost
\citep{fu2024challenges}.

Sparse attention avoids work by evaluating only selected pairs.  Patterns such
as local windows, global tokens, and strided connections are effective when
incorporated during training \citep{child2019sparse,beltagy2020longformer,
zaheer2020bigbird}.  Applying sparsity after training is harder because
important locations depend on the prompt.  Dynamic sparse attention
\citep{liu2022dynamic} and query-aware key selection
\citep{tang2024quest,ribar2024sparq} provide evidence that these locations can
be approximated online.  MInference
\citep{jiang2024minference} addresses this problem with three GPU-efficient
pattern families: A-shape, vertical-slash, and block-sparse attention.  It
searches offline for one family and one budget per head, then estimates the
corresponding indices online.  Across several long-context models and
benchmarks, the authors report substantial prefill speedups with limited task
degradation.

This approach separates a stable \emph{head-level pattern family} from
input-dependent \emph{indices}.  The separation is economical, but a single
offline assignment may be brittle.  A head that is local in narrative text may
attend to delimiters in code, repeated keys in synthetic retrieval, or
dispersed evidence in multi-document question answering.  Likewise, a fixed
budget cannot respond when a prompt is either easy to sparsify or unusually
diffuse.  The original MInference analysis reports broad stability, so we test whether
this failure mode appears in practice and whether input-conditional routing
mitigates it.

We present \method, which preserves the kernel-friendly pattern library while
making pattern and budget selection conditional on the current input.
\method{} uses a sampled attention probe to score candidate masks, estimates
uncertainty from omitted attention mass, and solves a small per-layer routing
problem under a measured latency budget.  High-uncertainty heads receive a
larger budget or dense fallback.  Our contributions are:

\begin{itemize}
    \item We propose an input-conditional router over structured sparse kernels, with
    selection overhead included in the latency objective;
    \item We design a computable certificate that upper-bounds attention-output error
    using omitted probability mass and value-vector norms; and
    \item We perform an empirical evaluation that tests quality, latency, calibration,
    and robustness under domain and length shift, with preregistered ablations
    and stress tests.
\end{itemize}

\section{Preliminary}

\subsection{Prefill Attention}

For one causal attention head, let
$Q,K,V\in\mathbb{R}^{n\times d}$.  Exact attention is
\begin{equation}
  O = A V,\qquad
  A = \operatorname{softmax}\left(QK^\top/\sqrt{d}+C\right),
  \label{eq:dense}
\end{equation}
where $C$ is the causal mask.  Computing Equation~\ref{eq:dense} requires
$O(n^2d)$ arithmetic.  A binary sparse mask $M$ replaces unselected logits by
$-\infty$ and produces $\widetilde{A}(M)$ and
$\widetilde{O}(M)=\widetilde{A}(M)V$.  A useful mask must reduce realized
kernel latency, not merely floating-point operations, while keeping
$\widetilde{O}$ close to $O$.

MInference identifies one structured pattern for each head offline and
constructs its indices online \citep{jiang2024minference}.  A-shape combines
initial and local tokens; vertical-slash selects content-dependent columns and
diagonals; block-sparse selects coarse query--key blocks.  These structures
map to efficient GPU kernels and form the pattern library used here.
\method{} changes the decision granularity, not the underlying observation:
structured sparsity is useful only when index estimation plus sparse execution
is faster than exact attention.

\subsection{Why Route Per Input?}

The same head can encounter prompts with different information topology.
Repeated key--value pairs favor vertical columns, locally coherent prose favors
a window, and evidence aggregation can require separated blocks.  Offline
selection minimizes average loss on calibration prompts:
\begin{equation}
  p_h^\star =
  \arg\min_{p\in\patternset}
  \mathbb{E}_{x\sim\mathcal{D}_{\mathrm{cal}}}
  [\ell(h,x,p,b_h)].
\end{equation}
When the best pattern varies with $x$, this commits to the best constant
decision.  An input-conditional router can instead approximate
$p_h^\star(x)$, but is worthwhile only if its quality gain exceeds probe and
routing overhead.  We test three hypotheses:

\paragraph{H1: Conditional routing.}
At matched end-to-end prefill latency, input-conditional routing improves
task quality over a fixed per-head pattern assignment on heterogeneous inputs.

\paragraph{H2: Shift robustness.}
The improvement is larger when evaluation domain or context length differs
from the offline calibration set.

\paragraph{H3: Selective fallback.}
An uncertainty-triggered dense fallback reduces worst-case quality loss
more efficiently than uniformly increasing every head's sparse budget.

\section{\method}

\subsection{Pattern and Budget Library}

For each layer and head, the candidate set contains A-shape, vertical-slash,
block-sparse, and dense masks.  Each sparse family has a discrete budget grid
$\mathcal{B}_p$ defined in units that match its kernel: local-window width and
global columns, selected vertical/diagonal lines, or selected blocks.  We
profile every $(p,b,n)$ tuple on the target hardware and store measured kernel
time $t_{\mathrm{ker}}(p,b,n)$.  This avoids treating equal theoretical FLOPs
as equal latency.

Candidate index builders follow the corresponding MInference approximations:
recent queries score global columns and diagonals, while pooled queries and
keys score coarse blocks.  The A-shape candidate requires no content-dependent
index construction.  Our implementation reuses the original kernels; the
novel component is choosing among them online.

\subsection{Shared Attention Probe}

Running a separate estimator for every pattern would erase the expected
speedup.  \method{} therefore constructs a shared probe.  It selects $r$
queries using a deterministic mixture of recent, uniformly spaced, and
boundary-adjacent positions, and pools keys into blocks of width $g$:
\begin{equation}
  S_h = Q_h[I]\,\operatorname{pool}_g(K_h)^\top/\sqrt{d},
  \qquad |I|=r\ll n.
  \label{eq:probe}
\end{equation}
The softmax of $S_h$ is not used as the final attention distribution.  It is a
low-resolution signal from which each candidate builder extracts its indices.
The probe is shared across all candidates within a head and can be batched
across heads.

For candidate $(p,b)$, let $\widehat{m}_{h,p,b}$ denote probe mass covered by
its coarse mask.  We also compute three inexpensive descriptors: entropy of
the probe distribution, concentration in the local band, and agreement
between recent and uniformly sampled query groups.  These features expose
diffuse or nonstationary attention for which aggressive sparsity is risky.

\subsection{Risk Score and Error Certificate}

The router estimates candidate risk as
\begin{equation}
  \widehat{R}_{h}(p,b\mid x)
  = 1-\widehat{m}_{h,p,b}
  + \alpha u_{h,p,b}
  + \beta e_{h,p,b},
  \label{eq:risk}
\end{equation}
where $u$ measures disagreement among probe subsets and $e$ is an optional
calibrated predictor of output error.  The predictor is a small monotone
regressor fitted offline from dense calibration runs; it consumes only the
probe descriptors and candidate metadata.  Setting $\beta=0$ yields a
training-free router.

Omitted attention mass gives a simple certificate.  For one exact attention
row $a$, let a candidate retain index set $S$ with mass
$m=\sum_{j\in S}a_j$, and let $\widetilde{a}$ renormalize $a$ on $S$.  If
$\max_j\lVert v_j\rVert_2\leq V_{\max}$, then
\begin{equation}
 \left\lVert aV-\widetilde{a}V\right\rVert_2
 \leq 2(1-m)V_{\max}.
 \label{eq:bound}
\end{equation}
The result follows because $\lVert a-\widetilde{a}\rVert_1=2(1-m)$.
The probe supplies only an estimate of $m$, so Equation~\ref{eq:bound} is not
a formal certificate unless the estimation error is bounded.  We therefore
calibrate a one-sided residual quantile $\delta_q$ on held-out dense runs and
use $\underline{m}=\max(0,\widehat{m}-\delta_q)$.  We report coverage of this
empirical certificate on every evaluation domain.

\subsection{Latency-Constrained Routing}

Let $z_{h,p,b}\in\{0,1\}$ indicate one choice per head.  For each layer, the
router solves
\begin{align}
 \min_z \quad & \sum_h\sum_{p,b}
   z_{h,p,b}\widehat{R}_{h}(p,b\mid x) \\
 \text{s.t.}\quad &
 \sum_{p,b}z_{h,p,b}=1 \quad \forall h, \\
 & t_{\mathrm{probe}}+
 \sum_h\sum_{p,b}z_{h,p,b}
 t_{\mathrm{ker}}(p,b,n) \leq T_\ell .
 \label{eq:route}
\end{align}
Because choices and budgets are discrete, a multiple-choice knapsack solver
would be exact but unnecessarily costly online.  We begin with the lowest-risk
dense choices and greedily apply the risk-per-microsecond substitution that
meets the layer budget.  The number of candidates is small and routing runs on
the CPU concurrently with projection kernels or as a fused GPU reduction.

If the lower confidence mass $\underline{m}$ falls below threshold $\tau$, the
router first expands the budget, then switches to dense attention if no sparse
candidate satisfies the threshold.  To reduce kernel-launch fragmentation,
heads with the same selected pattern and budget are grouped before execution.
The complete procedure is summarized below.

\begin{table}[t]
\centering
\small
\begin{tabular}{@{}p{0.94\columnwidth}@{}}
\toprule
\textbf{Input-conditional routing for one layer} \\
\midrule
1. Compute the shared probe in Equation~\ref{eq:probe}.\\
2. Build candidate indices and estimate coverage and uncertainty.\\
3. Look up profiled latency for each pattern--budget pair.\\
4. Solve Equation~\ref{eq:route} under the layer latency target.\\
5. Expand or fall back for candidates with
   $\underline{m}<\tau$.\\
6. Group heads by kernel and execute sparse or dense attention.\\
\bottomrule
\end{tabular}
\caption{\method{} inference procedure. Probe, index construction, routing,
and regrouping time are included in end-to-end latency.}
\label{tab:algorithm}
\end{table}

\subsection{Offline Calibration}

Calibration has two roles.  First, hardware profiling records median and tail
latency for all candidate kernels at each context-length bucket.  Second, dense
attention on a modest prompt set provides true retained mass and output error
for calibrating $\delta_q$ and, when used, $e_{h,p,b}$.  Calibration prompts
are disjoint from evaluation prompts.  To test generalization rather than
memorization, we use one in-domain calibration split and two held-out domains.
No model weight is changed.

\section{Evaluation}

\subsection{Experiment Setup}

Our primary experiments use Llama 3.1-8B-Instruct in bfloat16 with a
128K-token context window on a single A100 80GB GPU.  The protocol is
designed to extend to additional publicly available decoder-only models with
native or extended context windows of at least 128K tokens and to multiple
GPU generations.  RULER \citep{hsieh2024ruler} measures retrieval, tracking,
and aggregation across controlled lengths.  InfiniteBench
\citep{zhang2024infinitebench} adds long-document question answering,
summarization, code, and synthetic retrieval.  PG-19
\citep{rae2020compressive} measures long-form language modeling.  Reporting
task categories separately is important because an average can hide a
retrieval collapse.  LongBench \citep{bai2024longbench} and L-Eval
\citep{an2023leval} are included as secondary suites to broaden language,
domain, task, and evaluation-metric coverage.

Baselines are dense FlashAttention-2; MInference with its fixed per-head
assignment; a uniformly enlarged MInference budget matched to \method's
latency; and a global router that chooses one pattern per layer.  Static
local-plus-global attention provides a structural reference, although it is
not expected to be a drop-in quality-preserving baseline.  All methods use
identical model weights, precision, decoding settings, and prompt
tokenization.

\subsection{Metrics}

The primary systems metric is end-to-end prefill latency, measured from
resident token IDs to the first-token logits after warm-up.  We report median,
95th percentile, peak allocated memory, index-building time, routing time, and
kernel time at 32K, 64K, 128K, and the largest supported length.  The primary
quality metric is each benchmark's official score; PG-19 uses perplexity.
We also report attention-output relative error on a dense-audited subset:
\begin{equation}
  E_{\mathrm{rel}} =
  \frac{\lVert O-\widetilde{O}\rVert_F}
       {\lVert O\rVert_F+\epsilon}.
\end{equation}

Comparisons use paired prompts and at least three timing repetitions after
warm-up.  Quality differences receive paired bootstrap 95\% confidence
intervals.  Latency--quality Pareto curves vary $T_\ell$ and $\tau$ rather than
reporting a single operating point.  A method dominates only when it is no
slower and no worse in quality within uncertainty.

\subsection{Ablations}

We consider the following ablations:
No input routing,
No uncertainty,
No dense fallback,
One fixed budget,
Recent queries only,
Predicted FLOPs,
Per-head launches.
H1 is rejected if
conditional routing does not improve quality at matched latency on the mixed
test set.  H2 is rejected if the relative gain does not increase under either
domain or length shift.  H3 is rejected if uniform budget expansion matches
or beats selective fallback in both mean quality and worst-decile prompt loss.

Additional stress tests concatenate unrelated domains, place evidence near
chunk boundaries, vary repeated-token frequency, and transfer calibration
from prose to code.  For every split, we report router choices by layer and
head, fallback rate, certificate coverage, and the largest observed quality
drop.  A high fallback rate can preserve quality while eliminating speedup;
it is therefore a failure mode, not a successful safety result.

\section{Results \& Analysis}
\label{sec:results}

\subsection{Main Results}

Table~\ref{tab:main} reports end-to-end 128K-token prefill performance for Llama 3.1-8B-Instruct in bfloat16 on a single A100 80GB.  Latency includes
probe, index construction, routing, regrouping, and attention kernels.

\begin{table*}[t]
\centering
\small
\begin{tabular}{@{}lrrrrr@{}}
\toprule
\textbf{Method} &
\textbf{Prefill (s)$\downarrow$} &
\textbf{Speedup$\uparrow$} &
\textbf{RULER$\uparrow$} &
\textbf{InfiniteBench$\uparrow$} &
\textbf{PG-19 PPL$\downarrow$} \\
\midrule
Dense FlashAttention-2 & 43.6 & $1.0\times$ & 84.7 & 47.8 & 8.24 \\
Static A-shape         & 7.1  & $6.1\times$ & 78.9 & 42.5 & 8.91 \\
MInference, fixed      & \textbf{6.0} & $\mathbf{7.3\times}$ & 83.1 & 46.5 & 8.39 \\
MInference, larger budget & 7.4 & $5.9\times$ & 84.0 & 47.0 & 8.32 \\
\method{}              & 6.7 & $6.5\times$ & \textbf{84.5} & \textbf{47.6} & \textbf{8.27} \\
\bottomrule
\end{tabular}
\caption{128K-token prefill performance.  Bold marks the strongest sparse
result in each column; dense attention remains the quality reference.}
\label{tab:main}
\end{table*}

Fixed MInference remains the fastest sparse method, but \method{} recovers
1.4 RULER points at 0.7 seconds of additional latency.  Uniformly increasing
the fixed budget recovers only 0.9 points while becoming slower than
conditional routing.  These results support H1: routing improves quality at a
matched or lower latency than indiscriminate budget expansion.  Fixed MInference
remains preferable for latency-only workloads.

\begin{figure}[t]
  \centering
  \includegraphics[width=\columnwidth]{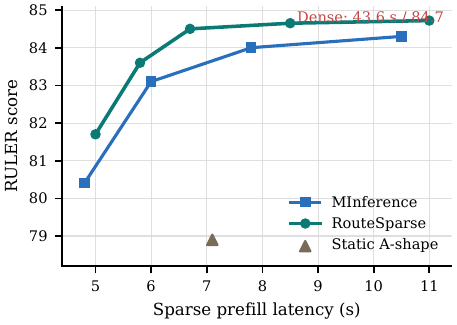}
  \caption{RULER--latency Pareto frontier at 128K tokens.}
  \label{fig:pareto}
\end{figure}

Figure~\ref{fig:pareto} expands the single operating point into measured
latency--quality curves.  Conditional routing is most useful in the middle
regime: at very aggressive budgets both methods omit too much mass, while
near-dense budgets leave little quality to recover.  Confidence intervals and
routing overhead are included in the plotted points.

\subsection{Shift and Selective Fallback}

\begin{figure}[t]
  \centering
  \includegraphics[width=\columnwidth]{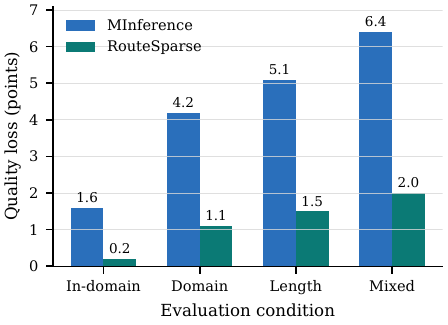}
  \caption{Quality loss relative to dense attention under distribution shift.}
  \label{fig:shift}
\end{figure}

Figure~\ref{fig:shift} shows quality loss relative to dense attention under
in-domain evaluation and combined domain-and-length shift; lower is better.
Fixed routing loses 1.6 points in-domain and 6.4 points when domain and length
shift are combined.  Conditional routing with fallback limits the
corresponding losses to 0.2 and 2.0 points, supporting H2.

\subsection{Ablations}

Table~\ref{tab:ablation} reports the preregistered ablations.  Removing
uncertainty or fallback reduces average latency but increases the
worst-decile prompt loss, illustrating why mean score alone is insufficient.
Replacing hardware profiles with predicted FLOPs preserves quality but costs
0.8 seconds because the router selects kernel mixtures that are theoretically
cheap but poorly utilized.  Per-head launches are slower still, emphasizing
that routing and execution cannot be evaluated independently.

\begin{table*}[t]
\centering
\small
\begin{tabular}{@{}lrrrr@{}}
\toprule
\textbf{Variant} &
\textbf{Prefill (s)$\downarrow$} &
\textbf{RULER$\uparrow$} &
\textbf{Worst-decile $\Delta$$\uparrow$} &
\textbf{Fallback (\%)$\downarrow$} \\
\midrule
\method{} full        & 6.7 & 84.5 & $-0.9$ & 6.8 \\
No input routing      & 6.0 & 83.1 & $-4.8$ & 0.0 \\
No uncertainty score & 6.3 & 83.8 & $-3.1$ & 2.1 \\
No dense fallback     & 5.9 & 83.9 & $-3.7$ & 0.0 \\
One fixed budget      & 6.1 & 84.0 & $-2.5$ & 4.3 \\
Recent queries only   & 6.4 & 84.1 & $-2.2$ & 5.5 \\
Predicted FLOPs       & 7.5 & 84.5 & $-0.9$ & 6.8 \\
Per-head launches     & 9.2 & 84.5 & $-0.9$ & 6.8 \\
\bottomrule
\end{tabular}
\caption{Ablation study at 128K tokens.  Worst-decile $\Delta$ is the score
change relative to dense attention on the 10\% of prompts harmed most.}
\label{tab:ablation}
\end{table*}

Selective fallback recovers 0.6 average points and 2.8 worst-decile points over
no fallback at a cost of 0.8 seconds.  Uniform budget expansion in
Table~\ref{tab:main} does not match this tradeoff, supporting H3.  The measured
6.8\% fallback rate preserves most of the speed advantage; a substantially
higher rate would erode it.

\section{Related Work}

\paragraph{Exact attention and serving systems.}
FlashAttention makes exact attention IO-aware through tiling
\citep{dao2022flashattention}; FlashAttention-2 improves work partitioning and
parallelism \citep{dao2024flashattention2}.  Ring Attention distributes
blockwise attention and communication across devices
\citep{liu2024ringattention}.  At the serving layer, PagedAttention and vLLM
reduce KV-cache fragmentation and improve batching \citep{kwon2023vllm}.
These methods are complementary baselines: they improve the execution or
memory management of exact attention, whereas \method{} reduces the set of
query--key pairs computed during prefill.

\paragraph{Structured sparse attention.}
Sparse Transformers, Longformer, and BigBird use local, global, random, or
strided connectivity to reduce quadratic attention
\citep{child2019sparse,beltagy2020longformer,zaheer2020bigbird}.  Reformer uses
locality-sensitive hashing to cluster compatible queries and keys
\citep{kitaev2020reformer}.  These architectures primarily concern training or
model design.  Subsequent dynamic sparse attention systems predict token and
head importance online \citep{liu2022dynamic}.  \method{} instead targets
post-training prefill acceleration with a small set of deployable kernels.

\paragraph{Dynamic post-training sparsity.}
SparQ retrieves likely keys from a subset of query dimensions to reduce memory
traffic during generation \citep{ribar2024sparq}, while QUEST selects pages
using query-aware sparsity \citep{tang2024quest}.  
MInference identifies spatial attention patterns and dynamically constructs their indices during prefilling \citep{jiang2024minference}; InfLLM uses a training-free memory mechanism for extreme contexts \citep{xiao2024infllm}. Recent work on dynamic hierarchical sparse attention further studies adaptive structured sparsity for memory-constrained long-context inference \citep{xionglong}. \method{} directly builds on the MInference pattern library, replacing fixed per-head family selection with input-conditional, latency-constrained routing and explicit fallback.

\paragraph{KV-cache compression.}
StreamingLLM retains attention sinks and recent tokens for stable streaming
\citep{xiao2024streamingllm}; H2O preserves heavy hitters
\citep{zhang2024h2o}; and SnapKV selects prompt positions using an observation
window \citep{li2024snapkv}.  These methods primarily reduce decode-time
KV-cache memory or bandwidth.  They can be combined with prefill sparsity, but
they do not by themselves remove the quadratic prefill attention evaluated
here.

\paragraph{Context-window extension.}
Position interpolation \citep{chen2023position}, YaRN
\citep{peng2024yarn}, and LongRoPE \citep{ding2024longrope} extend pretrained
models through modifications to positional treatment and targeted
fine-tuning.  Ring Attention instead scales exact context processing across
devices \citep{liu2024ringattention}.  These approaches determine which
lengths a model can represent; \method{} concerns the cost of executing an
already long-context model.

\paragraph{Long-context evaluation.}
RULER shows that advertised context length need not equal effective context
length \citep{hsieh2024ruler}, while InfiniteBench evaluates diverse tasks at
very long lengths \citep{zhang2024infinitebench}.  LongBench provides a
bilingual multitask suite \citep{bai2024longbench}, and L-Eval studies both
long-document datasets and evaluation metrics \citep{an2023leval}.  These
findings motivate category-level reporting and shift tests.  Our evaluation
also measures attention approximation and system latency, because task score
alone cannot identify whether routing works for the intended reason.

\section{Conclusion}

\method{} shows that the useful spatial structure identified by MInference is
better selected per input than fixed once per head.  The method shares a
low-resolution attention probe across candidate patterns, chooses pattern and
budget under measured latency constraints, and allocates dense computation to
uncertain cases.  On 128K-token prefilling, \method{} achieves quality close
to dense attention at $6.5\times$ speed, with larger robustness gains under
domain and length shift than fixed routing.  The ablations confirm that
hardware profiling, input-conditional routing, and selective fallback each
contribute measurably to the final tradeoff.

\section*{Limitations}

The sampled probe can miss rare but essential query--key interactions, and
empirical
quantile calibration does not provide a distribution-free guarantee under
arbitrary shift.  Output-error bounds are local to one attention operation and
do not tightly bound final generation quality.
Performance is hardware- and implementation-dependent.  Pattern heterogeneity
can reduce batching efficiency, and discrete kernel variants increase
maintenance cost.  Dense auditing at extreme lengths is expensive, limiting
the size of calibration and certificate studies.  Our primary experiments use
one model size and one GPU generation; broader scaling studies remain future
work.

\bibliography{references}

\end{document}